\documentclass[letterpaper, 10 pt, conference]{ieeeconf}  

\IEEEoverridecommandlockouts                              

\usepackage{epsfig} 
\usepackage{amsmath} 
\usepackage{amssymb}  
\usepackage{multirow}

\usepackage{hyperref}       
\usepackage{url}            
\usepackage{booktabs}       

\usepackage{nicefrac}       
\usepackage{microtype}      
\usepackage{xcolor}         

\usepackage[scientific-notation=true]{siunitx}

\usepackage{microtype}
\usepackage{graphicx}

\usepackage{amsfonts}       
\usepackage{amsmath}
\usepackage{bbm}

\usepackage{algorithm}
\usepackage{algorithmic}

\let\labelindent\relax
\usepackage{enumitem}

\usepackage{soul}

\usepackage{cuted}

\newcommand{\cS}{\mathcal{S}}
\newcommand{\cP}{\mathcal{P}}
\newcommand{\cA}{\mathcal{A}}
\newcommand{\cR}{\mathcal{R}}
\newcommand{\cD}{\mathcal{D}}

\newcommand{\cH}{\mathcal{H}} 
\newcommand{\cL}{\mathcal{L}} 

\newcommand{\stt}{s_t}
\newcommand{\at}{a_t}
\newcommand{\rt}{r_t}
\newcommand{\stone}{s_{t+1}}
\newcommand{\atone}{a_{t+1}}

\newcommand{\sa}{(s, a)}

\DeclareMathOperator*{\E}{\mathbb{E}} 
\newcommand{\idc}[1]{1\big[#1\big]}  

\DeclareMathOperator*{\argmin}{\arg\,min}

\newcommand{\bmax}{\beta_{max}}
\newcommand{\bmin}{\beta_{min}}

\newcommand{\qhat}{\hat{Q}}
\newcommand{\qhatpi}{\hat{Q}^{\pi}}
\newcommand{\qpi}{Q^{\pi}}
\newcommand{\qbeta}{Q_{\beta}}
\newcommand{\qhatbeta}{\hat{Q}_{\beta}}

\newcommand{\pluseq}{\mathrel{+}=}

\newtheorem{theorem}{Theorem}

\title{\LARGE \bf
Adaptively Calibrated Critic Estimates for Deep Reinforcement Learning
}

\author{Nicolai Dorka$^{1}$ and Tim Welschehold$^{1}$ and Joschka Bödecker$^{1}$ and Wolfram Burgard$^{2}$

\thanks{Authors are with: $^{1}$University of Freiburg and $^{2}$ University of Technology Nuremberg, Germany.
        {\tt\small dorka@cs.uni-freiburg.de}.}%
\thanks{This work was supported by the European Union’s Horizon 2020 research and innovation program under grant agreement No 871449-OpenDR.}%
        }

\begin{document}

\maketitle
\thispagestyle{empty}
\pagestyle{empty}

\begin{abstract}

Accurate value estimates are important for off-policy reinforcement learning. 
Algorithms based on temporal difference learning typically are prone to an over- or underestimation bias building up over time.
In this paper, we propose a general method called Adaptively Calibrated Critics (ACC) that uses the most recent high variance but unbiased on-policy rollouts to alleviate the bias of the low variance temporal difference targets.
We apply ACC to Truncated Quantile Critics~\cite{tqc}, which is an algorithm for continuous control that allows regulation of the bias with a hyperparameter tuned per environment.
The resulting algorithm adaptively adjusts the parameter during training rendering hyperparameter search unnecessary 
and sets a new state of the art on the OpenAI gym continuous control benchmark among all algorithms
that do not tune hyperparameters for each environment.
ACC further achieves improved results on different tasks from the Meta-World robot benchmark.
Additionally, we demonstrate the generality of ACC by applying it to TD3~\cite{td3} and showing an improved performance also in this setting. 
\end{abstract}


\section{Introduction}

Off-policy reinforcement learning is an important research direction as the reuse of old experience promises to make these methods more sample efficient than their on-policy counterparts. This is an important property for many applications such as robotics where  interactions with the environment are very time- and cost-intensive.
Many successful off-policy methods make use of a learned Q-value function~\cite{td3,SAC,hessel2018rainbow,dqn15}. 
If the action space is discrete the Q-function can be directly  used to generate actions while for continuous action spaces it is usually used in an actor-critic setting where the policy is trained to choose actions that maximize the Q-function. In both cases accurate estimates of the Q-values are of crucial importance.

Unfortunately, learning the Q-function off-policy can lead to an overestimation bias~\cite{Thrun+Schwartz:1993}.
Especially when a nonlinear function approximator is used to model the Q-function, there are many potential sources of bias.
Different heuristics were proposed for their mitigation, such as the double estimator in the case of discrete action spaces~\cite{hasselt2016deepdouble} or taking the minimum of two estimates in the case of continuous actions~\cite{td3}.
While these methods successfully prevent extreme overestimation, due to their coarse nature, they can  still induce under- or overestimation bias to a varying degree depending on the environment~\cite{Lan2020Maxmin}.\looseness=-1

To overcome these problems we propose a principled and general method to alleviate the bias called Adaptively Calibrated Critics (ACC).
Our algorithm uses the most recent on-policy rollouts to determine the current bias of the Q-estimates and adjusts a bias controlling parameter accordingly.
This parameter adapts the size of the temporal difference (TD) targets  such that the bias can be corrected in the subsequent updates.
As the parameter changes slower than the rollout returns, our method still benefits from stable and low-variance temporal difference targets, while it incorporates the information from unbiased but high variance samples from the recent policy to reduce the bias.

{\spaceskip= 2pt plus 1pt minus 1.5pt  \spaceskip= 3pt plus 2pt minus 2pt We apply ACC to Truncated Quantile Critics (TQC) \cite{tqc}, which is a recent off-policy actor-critic algorithm for continuous control showing strong performance on various tasks. 
In TQC the bias can be controlled in a finegrained way with the help of a hyperparameter that has to be tuned for every environment.
ACC allows to automatically adjusts this parameter online during the training in the environment.
As a result, it eliminates the need to tune this hyperparameter in a new environment, which is very expensive or even infeasible for many applications.}

We evaluate our algorithm on a range of continuous control tasks from OpenAI gym \cite{gymopenai} and robotic tasks from the meta world benchmark \cite{yu2020meta} and exceed the current state-of-the-art results among all algorithms that do not need  tuning of environment-specific hyperparameters.
For each environment, ACC matches the performance of TQC with the optimal hyperparameter for that environment.
Further, we show that the automatic bias correction allows to increase the number of value function updates performed per environment step, which results in even larger performance gains in the sample-efficient regime.
We additionally apply ACC to the TD3 algorithm \cite{td3} where it also leads to notably improved performance, underscoring the generality of our proposed method.
To summarize, the main contributions of this work are:
\begin{enumerate}[leftmargin=0.72cm]
    \item We propose Adaptively Calibrated Critics, a new general algorithm  that reduces the bias of value estimates in a principled fashion with the help of the most recent unbiased on-policy rollouts.
    \item As a practical implementation we describe how ACC can be applied to learn a bias-controlling hyperparameter of the TQC algorithm and show that the resulting algorithm sets a new state of the art on the OpenAI continuous control benchmark suite.
    \item ACC achieves strong performance on robotics tasks.
    \item We demonstrate that ACC is a general algorithm with respect to the adjusted parameter by additionally applying it successfully to TD3.
\end{enumerate}
\looseness=-1


To allow for reproducibility of our results we describe our algorithm in detail, report all hyperparameters, use a large number of random seeds for evaluation, and made the source code publicly available\footnote{\url{https://github.com/Nicolinho/ACC}}.

\section{Background}

We consider model-free reinforcement learning for episodic tasks with continuous state and action spaces $\cS$ and $\cA$. An agent interacts with its environment by selecting an action $\at \in \cA$ in state $\stt \in \cS$ for every discrete time step $t$. The agent receives a scalar reward $r_t$ and transitions to a new state $s_{t+1}$.
To model this in a mathematical framework we use a Markov decision process, defined by the tuple 
$(\cS, \cA, \cP, \cR, \gamma)$. Given an action $a \in \cA$ in state $s \in \cS$ the unknown state transition density $\cP$ defines a distribution over the next state.
Rewards come from the reward function $\cR$ and future rewards are discounted via the discount factor $\gamma \in [0, 1]$.

The goal is to learn a policy $\pi$ mapping a state $s$ to a distribution over actions such that the sum of future discounted rewards $R_t = \sum_{i=t}^{T} \gamma^{i-t} r_i$ is maximized.
We use the term $\pi_\phi$ for the policy with parameters $\phi$ trained to maximize the expected return 
$J(\phi) = \E_{s_i \sim \cP , a_i \sim \pi} [R_0]$.
The value function for a given state-action pair $(s,a)$ is defined as 
$Q^\pi (s,a) = \E_{s_i \sim \cP , a_i \sim \pi} [ R_t |s,a]$, which is the expected return when executing action $a$ in state $s$ and following $\pi$ afterwards.

\subsection{Soft Actor Critic}
TQC extends Soft Actor-Critic (SAC) \cite{SAC}, which is a strong off-policy algorithm for continuous control using entropy regularization.
While in the end we are interested in maximizing the performance with respect to the total amount of reward collected in the environment, SAC maximizes for an auxiliary objective that augments the original reward with the entropy of the policy
$J(\phi) = \E_{\stt \sim \cP , \at \sim \pi} [ \sum_t \gamma^{t} (r_t + \alpha \cH (\pi(\cdot | \stt)))   ]$, where $\cH$ denotes the entropy. 

A critic is learned that evaluates the policy $\pi$ in terms of its Q-value of the entropy augmented reward.
The policy---called actor---is trained to choose actions such that the Q-function is maximized with an additional entropy regularization
\vspace{-0.1cm}
\begin{equation}
    J_\pi (\phi) = \E_{\stt \sim \cD, \at \sim \pi_\phi} 
                    [ Q_\theta(\stt, \at) - \alpha \log \pi_\phi (\at|\stt) ].
\vspace{-0.1cm}
\end{equation}
The weighting parameter $\alpha$ of the entropy term can be automatically adjusted during the training~\cite{SACalgapp}.
Both the training of actor and critic happen off-policy with transitions sampled from a replay buffer.

\subsection{Truncated Quantile Critics}

The TQC algorithm uses distributional reinforcement learning \cite{bellemare2017distributional} to learn
a distribution over the future augmented reward instead of a Q-function which is a point estimate for the expectation of this quantity.
To do so TQC utilizes quantile regression \cite{dabney2018distributional} to approximate the distribution with Dirac delta functions
$Z_\theta (\stt,\at) = \frac{1}{M} \sum_{m=1}^{M} \delta ( \theta^m (\stt,\at))$.
The Diracs are located at the quantile locations for fractions 
$\tau_m = \frac{2m -1}{m}, m \in \{1, \dots, M \}$. The network is trained to learn the quantile locations $\theta^m (s,a)$  
by regressing the predictions $\theta^m (\stt, \at)$ onto the Bellman targets 
$y_m(\stt, \at) = \rt + \gamma ( \theta^m (\stone, \atone) - \alpha \log \pi_\phi (\atone | \stone ))$ 
via the  Huber quantile loss.

TQC uses an ensemble of $N$ networks $(\theta_1, \cdots, \theta_N)$  where each network $\theta_n$ predicts the distribution 
$Z_{\theta_n} (\stt,\at) = \frac{1}{M} \sum_{m=1}^{M} \delta ( \theta_n^m(\stt,\at))$.
A single Bellman target distribution is computed for all networks. This happens by first computing all targets for all networks, pooling all targets together in one set and sorting them in ascending order. 
Let $k \in \{1, \dots, M\}$, then the $kN$ smallest of these targets $y_i$ are used to define the target distribution
$Y(\stt, \at) = \frac{1}{kN} \sum_{i=1}^{kN} \delta ( y_i (\stt,\at))$.
The networks are trained by minimizing the quantile Huber loss which in this case is given by
\vspace{-0.2cm}
\begin{equation}
    L(\stt,\at; \theta_n) \hspace{-1pt} = \hspace{-1pt} \frac{1}{kNM} \hspace{-2pt} \sum_{m,i=1}^{M, kN} \hspace{-2pt} \rho^H_{\tau_m} \hspace{-1pt} (y_i(\stt,\at)  -  \theta^m_n (\stt,\at))
    \vspace{-0.2cm}
\end{equation}
where $\rho^H_{\tau} (u) = |\tau -  \mathbf{1}(u < 0) | \cL_H^1(u)$ and $\cL_H^1(u)$ is the Huber loss with parameter $1$.

The rationale behind truncating some quantiles from the target distribution is to prevent overestimation bias. 
In TQC the number of dropped targets per network $d = M- k$ is a hyperparameter that has to be tuned per environment but allows for a finegrained control of the bias.

The policy is trained as in SAC by maximizing the entropy penalized estimate of the Q-value which is the expectation over the distribution obtained from the critic
\begin{equation}
J(\phi) = \E_{\substack{s\sim\cD\\ a\sim\pi}} \Bigg[ \frac{1}{NM} \sum_{m,n=1}^{M,N} \theta_n^m (s,a)  - \alpha \log \pi_\phi(a|s)   \Bigg]   .
\end{equation}

\section{Adaptively Calibrated Critics }

In this section, we will introduce the problem of estimation bias in TD learning, present our method ACC and demonstrate how it can be applied to TQC.

\subsection{Over- and Underestimation Bias}

The problem of overestimation bias in temporal difference learning with function approximation has been known for a long time \cite{Thrun+Schwartz:1993}.
In Q-learning \cite{watkins1992q} the predicted Q-value $Q(\stt,\at)$ is regressed onto the target given by $y = \rt + \gamma \max_a Q(\stone, a)$.
In the tabular case and under mild assumptions the Q-values converge to that of the optimal policy \cite{watkins1992q} with this update rule. However, using a function approximator to generate the Q-value introduces an approximation error.
Even under the assumption of zero mean noise corruption of the Q-value
$\E[\epsilon_a] = 0$,  an overestimation bias occurs in the computation of the target value because of Jensen's inequality
\vspace{-0.2cm}
\begin{align}
    \max_a Q(\stone, a) &= \max_a \E [ Q(\stone, a) + \epsilon_a] \nonumber \\
    & \leq \E \big[\max_a \{Q(\stone, a) + \epsilon_a\} \big] .
    \vspace{-0.9cm}
\end{align}\\[-0.5cm]
In continuous action spaces it is impossible to take the maximum over all actions. The most successful algorithms rely on an actor-critic structure where the actor is trained to choose actions that maximize the Q-value \cite{td3,SAC,ddpg}. So the actor can be interpreted an approximation to the argmax of the Q-value.

With deep neural networks as function approximators other problems such as over-generalization \cite{dqn15,NIPS17-ishand} can occur where the updates to $Q(\stt,\at)$ also increases the target through $Q(\stone,a)$ for all $a$ which could lead to divergence.
There are many other potential sources for overestimation bias such as  stochasticity of the environment 
 \cite{hasselt2010double} or computing the Q-target from actions that lie outside of the current training data distribution \cite{kumarStabilizing19}.

While for discrete action spaces the overestimation can be controlled with the double estimator \cite{hasselt2016deepdouble,hasselt2010double}, it was shown that this estimator does not prevent overestimation when the action space is continuous \cite{td3}. 
As a solution the TD3 algorithm \cite{td3} uses the minimum of two separate estimators to compute the critic target. This approach was shown to prevent overestimation but can introduce an underestimation bias.
In TQC \cite{tqc} the problem is handled by dropping some targets from the pooled set of all targets of an ensemble of distributional critics. This allows for more finegrained control of over- or underestimation by choosing how many targets are dropped. 
TQC is able to achieve an impressive performance but the parameter $d$ determining the number of dropped targets has to be set for each environment individually. This is highly undesirable for many applications
since the hyperparameter sweep to determine a good choice of the parameter increases the actual number of environment interactions proportional to the number of hyperparameters tested. For many applications like robotics this makes the training prohibitively expensive.

\subsection{Dynamically Adjusting the Bias}
In the following we present a new general approach to adaptively control bias emerging in TD targets regardless of the source of the bias.
Let $R^\pi \sa$ be the random variable denoting the sum of future discounted rewards when the agent starts in state $s$, executes action $a$ and follows policy $\pi$ afterwards. This means that the Q-value is defined as its expectation $\qpi\sa = \E[R^\pi \sa]$. For notational convenience we will drop the dependency on the policy $\pi$ in the following.
We start with the tabular case. Suppose for each state-action pair $\sa$ we have a family $\{\qhat_\beta \sa \}_{\beta \in [\bmin , \bmax] \subset \mathbb{R}}$ of estimators for $Q\sa$ with the property that
$\qhat_{\bmin}(s,a) \leq Q\sa \leq  \qhat_{\bmax}(s,a)$, where $Q\sa$ is the true Q-value of the policy $\pi$ and $Q_\beta$ a continuous monotone increasing function in $\beta$ .

If we have samples $R_i \sa$ of the discounted returns $R \sa$, an unbiased estimator for $Q\sa$ is given by the average of the $R_i$ through Monte Carlo estimation \cite{introdrl2018}.
We further define the estimator $\qhat_{\beta^*}\sa$, where $\beta^*$ is given by
\begin{equation}
    \beta^* \sa = \argmin_{\beta \in [\bmin, \bmax]} \Bigg| \qhat_\beta \sa - \frac{1}{N} \sum_{i=1}^{N} R_i \sa  \Bigg| .
    \label{eq:optimal_q_estimator}
\end{equation}

The following Theorem, which we prove in the appendix, shows that the estimator is unbiased under some assumptions.
\begin{theorem}
Let $Q_\beta \sa$ be a continuous monotone increasing function in $\beta$ and
 assume that for all $\sa$ it holds $\qhat_{\bmin}(s,a) \leq Q\sa \leq  \qhat_{\bmax}(s,a)$, the returns $R\sa$ follow a symmetric probability distribution and that $\qhat_{\bmin}(s,a)$ and $\qhat_{\bmax}(s,a)$ have the same distance to $Q\sa$.
Then $Q_{\beta^*}$ from Equation \ref{eq:optimal_q_estimator} is an unbiased estimator for the true value $Q$ for all $\sa$.
\end{theorem}
The symmetry and same distance assumption can  be replaced by assuming that $\qhat_{\bmin}(s,a)  \leq R_i \leq \qhat_{\bmax}(s,a) $ with probability one. In this case the proof is straightforward since $\qbeta$ can take any value for which $R_i$ has positive mass. 

\begin{figure}
\begin{algorithm}[H]
   \caption{ACC - General}
   \label{alg:general_acc}
\begin{algorithmic}
   \STATE {\bfseries Initialize:} bias controlling parameter $\beta$, steps between $\beta$ updates $T_\beta$, $t_\beta = 0$
   \FOR{$t=1$ {\bfseries to} total number of environment steps}
   \STATE Interact with environment according to $\pi$, store transitions in replay buffer $\mathcal{B}$ and store observed returns $R\sa$, increment $t_\beta \pluseq 1$
   \IF{episode ended \textbf{and} $t_\beta >= T_\beta$}
   \STATE Update $\beta$ with Eq. \ref{eq:one_step_beta_update} using the most recent experience and set $t_\beta=0$
   \ENDIF
   \STATE Sample mini-batch $b$ from $\mathcal{B}$
   \STATE Update $Q$ with target computed from $\qbeta$ and $b$
  \ENDFOR
\end{algorithmic}
\end{algorithm}
\vspace{-0.8cm}
\end{figure}

We are interested in the case where $\qhat$ is given by a function approximator such that there is generalization between state-action pairs and that it is possible to generate estimates for pairs for which there are no samples of the return available.
Consider off-policy TD learning where the samples for updates of the Q-function are sampled from a replay buffer of past experience.
While the above assumptions might not hold anymore in this case, we have an estimator for all state-action pairs and not just the ones for which we have samples of the return.
Also in practice rolling out the policy several times from each state action pair is undesirable and so we set $N=1$ which allows the use of the actual exploration rollouts.
Our proposed algorithm starts by initializing the bias-controlling parameter $\beta$ to some value.
After a number of environment steps and when the next episode is finished, the Q-value estimates and actual observed returns are compared. Depending on the difference $\beta$ is adjusted according to  
\vspace{-0.2cm}
\begin{equation}
    \beta_{new} = \beta_{old} + \alpha \sum_{t=1}^{T_\beta} \Big[   R (s_t, a_t) - \hat{Q} (s_t, a_t) \Big], 
    \label{eq:one_step_beta_update}
\vspace{-0.2cm}
\end{equation}
where $\alpha$ is a step size parameter and $(s_t, a_t)_{t=1}^{T_\beta}$ are the $T_\beta \in \mathbb{N}$ most recent state-action pairs.
As a result $\beta$ is decreased in the case of overestimation, where the Q-estimates are larger than the actual observed returns, 
and increased in the case of underestimation. 
We assumed that $\qbeta$ is continuous and monotonically increasing in $\beta$.  Hence, increasing $\beta$ increases $\qbeta$ and vice versa.
For updating the Q-function the target will be computed from $\qbeta$.

Only performing one update step and not the complete minimization from Equation \ref{eq:optimal_q_estimator} has the advantage that $\beta$ is changing relatively slow which means the targets are more stable.
Through this mechanism our method can incorporate the high variance on-policy samples to correct for under- or overestimation bias. 
At the same time our method can benefit from the low variance TD targets.
ACC in its general form is summarized  in Algorithm \ref{alg:general_acc}.

Other algorithms that attempt to control the bias arising in TD learning with non-linear function approximators usually use some kind of heuristic that includes more than one estimator.
Some approaches use them to decouple the choice of the maximizing action and the evaluation of the maximum in the computation of the TD targets \cite{hasselt2016deepdouble}. 
Alternative approaches take the  minimum, maximum or a combination of both over the different estimators \cite{td3,Lan2020Maxmin,agarwal2020optimistic,fujimoto2019off}.
All of these have in common that the same level of bias correction is done for every environment and for all time steps during training.
In the deep case there are many different sources that can influence the tendency of TD learning building up bias in non-trivial ways.
ACC is more principled in the regard that it allows to dynamically adjust the magnitude and direction of bias correction during training.
Regardless of the source and amount of bias ACC provides a way to alleviate it. This makes ACC promising to work robustly on a wide range of different environments.

One assumption of ACC is that there is a way to adjust the estimated Q-value with a parameter $\beta$ such that $\qhatbeta$ is continuous and monotonically increasing in $\beta$. 
There are many different functions that are in accordance with this assumption.
We give one general example of how such a $\qhatbeta$ can be easily constructed for any algorithm that learns a Q-value.
Let $\qhat$ be the current estimate. 
Then define $\qhatbeta = \beta |\qhat| / K + \qhat$, where $K$ is a constant (e.g. $100$) and $[\bmin,\bmax]$ is some interval around $0$.
In the following section we will present an application of ACC in a more sophisticated way.

\subsection{ Applying ACC to TQC}

As a practical instantiation of the general ACC algorithm we apply it to adjust the number of targets dropped from the set of all targets in TQC. 
Denote with $d_{max} \in \{0,\dots, M \}$ some upper limit of targets to drop per network.
Define $\bmin=0$, $\bmax=d_{max}$ and let $d = d_{max} - \beta$ be the current number of targets dropped for each network. Further, we write $\qbeta$ for the TQC estimate with $dN$ targets dropped from the pooled set of all targets. 
If $d_{max}$ is set high enough the TQC estimate without dropped targets $Q_{\bmax}$ induces overestimation 
while the TQC estimate with $d_{max}$ dropped targets per net $Q_{\bmin}$ induces underestimation. 

In general, $\beta \in [0,d_{max}]$ is continuous and hence also $d$ is a continuous value. As  the number of dropped targets from the pooled set of all targets has to be a discrete number in $\{0, \dots, NM\}$ we round the total number of dropped targets $d N$ to the nearest integer in the computation of the TD target.
When updating $\beta$ with Equation \ref{eq:one_step_beta_update}, we divide the expectation by the moving average of the absolute value of the difference between returns and estimated Q-values for normalization.
\section{Experiments}

We evaluate our algorithm on a range of continuous control tasks from OpenAI Gym \cite{gymopenai} and the meta world benchmark \cite{yu2020meta} that both use  the physics engine MuJoCo \cite{mujoco} (version 1.5). 
First, we benchmark ACC against strong methods that do not use environment specific hyerparameters.
Then we compare the performance of TQC with a fixed number of dropped targets per network with that of ACC.
Next, we evaluate the effect of more critic updates for ACC and show results in the sample efficient regime.
Further, we study the effect of ACC on the accuracy of the value estimate, and investigate the generality of ACC by applying it to TD3.

We implemented ACC on top of the PyTorch code published by the authors\footnote{\url{https://github.com/bayesgroup/tqc_pytorch}} to ensure a fair comparison.
While in general a safe strategy is to use a very high value for $d_{max}$ as it gives ACC more flexibility in choosing the right amount of bias correction we set it to $d_{max}=5$, which is the maximum value used by TQC for the number of dropped targets in the original publication.
At the beginning of the training we initialize $\beta = 2.5$ and set the step size parameter to $\alpha=0.1$.
After $T_\beta = 1000$ steps since the last update and when the next episode finishes, $\beta$ is updated with a batch that stores the most recent state-action pairs encountered in the environment and their corresponding observed discounted returns. 
After every update of $\beta$ the oldest episodes in this stored batch are removed until there are no more than $5000$ state-action pairs left.
This means that on average $\beta$ is updated with a batch whose size is a bit over $5000$. 
The updates of $\beta$ are started after $25000$ environment steps and
the moving average parameter in the normalization of the $\beta-$update is set to $0.05$. 
The  first $5000$ environment interactions are generated with a random policy after which learning starts.
We did not tune most of these additional hyperparameters and some choices are directly motivated by the environment (e.g. setting $T_\beta$ to the maximum episode length). Only for $\alpha$ we tested a few different choices but found that for reasonable values it does not have a noticeable influence on performance. 
All hyperparameters of the underlying TQC algorithm  with $N=5$ critic networks were left unchanged.

Compared to TQC the additional computational overhead caused by ACC is minimal because there is only one update to $\beta$ that is very cheap compared to one training step of the actor-critic and there are at least $T_\beta =1000$ training steps in between one update to $\beta$.

During training, the policy is evaluated every 1,000 environment steps by averaging the episode returns of $10$ rollouts with the current policy. For each task and algorithm we run 10 trials each with a different random seed.

\subsection{Comparative Evaluation}

We compare ACC to the state of the art continuous control methods SAC \cite{SAC} (with learned temperature parameter \cite{SACalgapp}) and TD3 \cite{td3} on six OpenAI Gym continuous control environments.
To make the different environments comparable we normalize the scores by dividing the achieved return by the best achieved return among all evaluations points of all algorithms for that environment.

Figure \ref{fig:comparative_aggregated_results}a)  shows the aggregated data efficiency curve over all $6$ tasks computed with the method of \cite{agarwal2021deep}, where the interquantile mean (IQM) ignores the bottom and top $25$\% of the runs across all games and computes the mean over the remaining. 
The absolute performance of ACC for each single task can be seen in Figure \ref{fig:ablation_const_number_dropped_atoms_single_curves}.
Overall, ACC reaches a much higer performance than SAC and TD3.

\subsection{Robotics Benchmark}
\begin{figure}[t]
\footnotesize
\setlength{\tabcolsep}{1pt}
\centering 
\begin{tabular}{cc}
        \includegraphics[width=.49\linewidth]{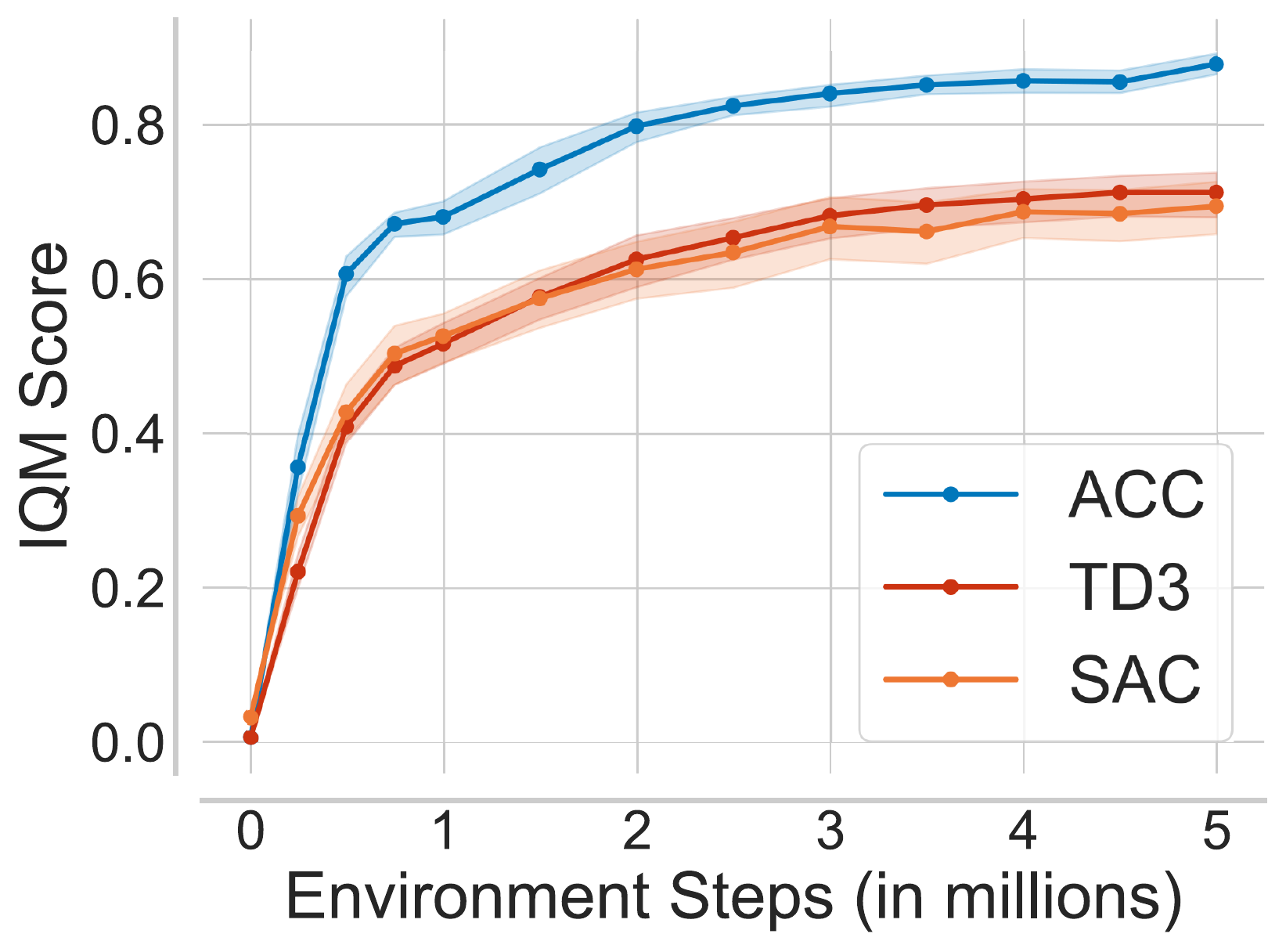} &
        \includegraphics[width= .49\linewidth]{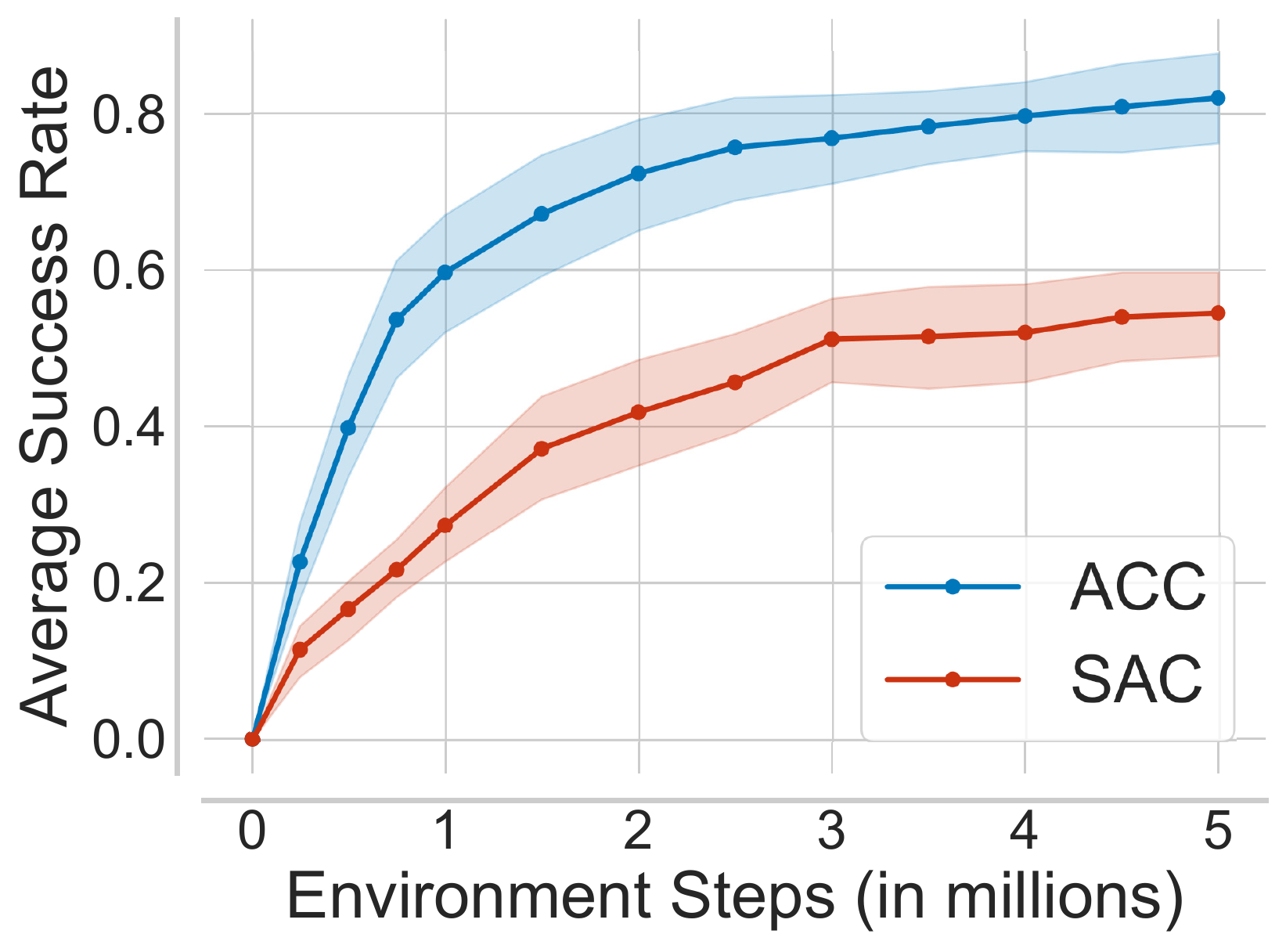} \\
        a) & b) \\
\end{tabular}
\vspace{-0.3cm}
\caption{
Sample efficiency curves aggregated from the results over several environments. The normalized IQM score and the mean of the success rate respectively is plotted against the number of environment steps. Shaded regions denote pointwise $95$\% stratified bootstrap confidence intervals according to the method of \cite{agarwal2021deep}. 
\textbf{(a)} Aggregated results over the $6$ gym continuous control tasks.
\textbf{(b)} Aggregated results over the $12$ metaworld tasks.
}
\label{fig:comparative_aggregated_results}
\vspace{-0.5cm}
\end{figure}

To investigate, if ACCs strong performance also translates into robotics environments, we evaluate ACC and SAC on $12$ of the more challenging tasks in the Meta-World benchmark \cite{yu2020meta}, which consists of several manipulation tasks with a Sawyer arm. We use version V2 and use the following $12$ tasks:
sweep, stick-pull, dial-turn, door-open, peg-insert-side, push, pick-out-of-hole, push-wall, faucet-open, hammer, stick-push, soccer.
We evaluate the single tasks in the in the MT1 version of the benchmark, where the goal and object positions change across episodes.
Different to the gym environments, $\beta$ is updated every $500$ environment steps as this is the episode length for these tasks.
Figure 
\ref{fig:comparative_aggregated_results}b)
shows the aggregated data efficiency curve in terms of success rate over all $12$ tasks computed with the method of \cite{agarwal2021deep}.

The curves demonstrate that ACC achieves drastically stronger results than SAC both in terms of data efficiency and asymptotic performance.
After $2$ million steps ACC already achieves a close to optimal task success rate which is even considerably higher than what SAC achieves at the end of the training.
This shows, that ACC is a promising approach for real world robotics applications.

\subsection{Fixing the Number of Dropped Targets}

\begin{figure}
    \centering
    \includegraphics[width=0.93\linewidth]{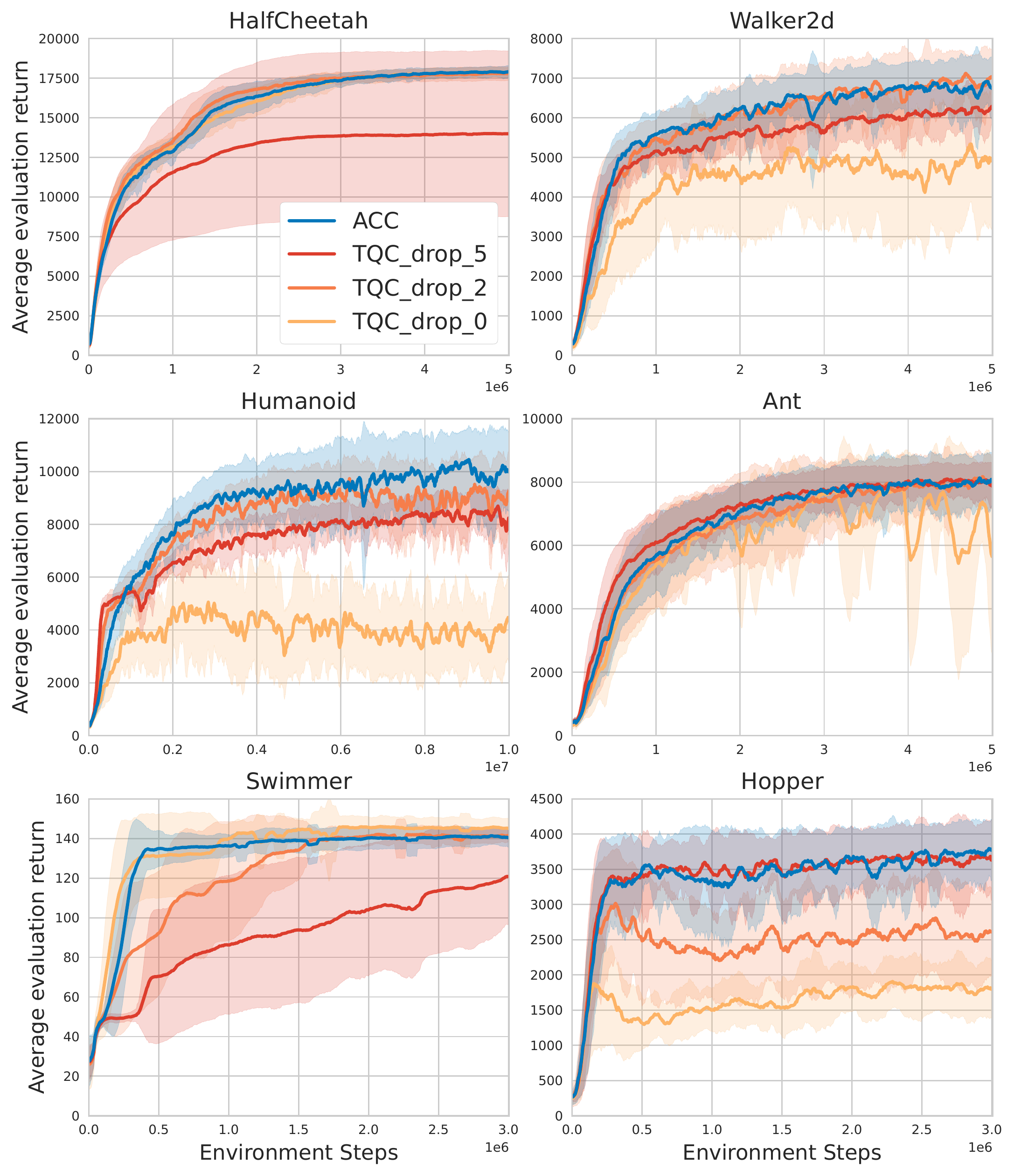}
    \caption{Learning curves of ACC applied to TQC and TQC with different fixed choices for the number of dropped atoms $d$ on six OpenAi gym environments. We used version \textit{v3}. The shaded area represents  mean $\pm$ standard deviation over the $10$ trials. For readability the curves showing the mean are filtered  with a uniform filter of size $15$.}
    \label{fig:ablation_const_number_dropped_atoms_single_curves}
\vspace{-0.5cm}
\end{figure}

In this experiment we evaluate how well ACC performs when compared to TQC where the number of dropped targets per network $d$ is fixed to some value.
Since in the original publication for each environment the optimal value was one of the three values $0$, $2$, and $5$, we evaluated TQC with $d$ fixed to one of these values for each environment.
To ensure comparability we used the same codebase as for ACC. 
The results in Figure \ref{fig:ablation_const_number_dropped_atoms_single_curves} show that it is not possible to find one value for $d$ that performs well on all environments.
With $d=0$, TQC is substantially worse on three environments and unstable on the \textit{Ant} environment.
Setting $d=2$ is overall the best choice but still performs clearly worse for two environments and is also slightly worse for \textit{Humanoid}.
Dropping $d=5$ targets per network leads to an algorithm that can compete with ACC only on two of the six environments.
Furthermore, even if there would be one tuned parameter that performs equally well as ACC on a given set of environments we hypothesize there are likely very different environments for which the specific parameter choice will not perform well. The principled nature of ACC on the other hand provides reason to believe that it can perform robustly on a wide range of different environments. This is supported by the robust performance on all considered environments.

\subsection{Evaluation of Sample Efficient Variant}

\begin{figure*}[t]
\footnotesize
\centering 
\begin{tabular}{cc}
    \includegraphics[width=.56\linewidth]{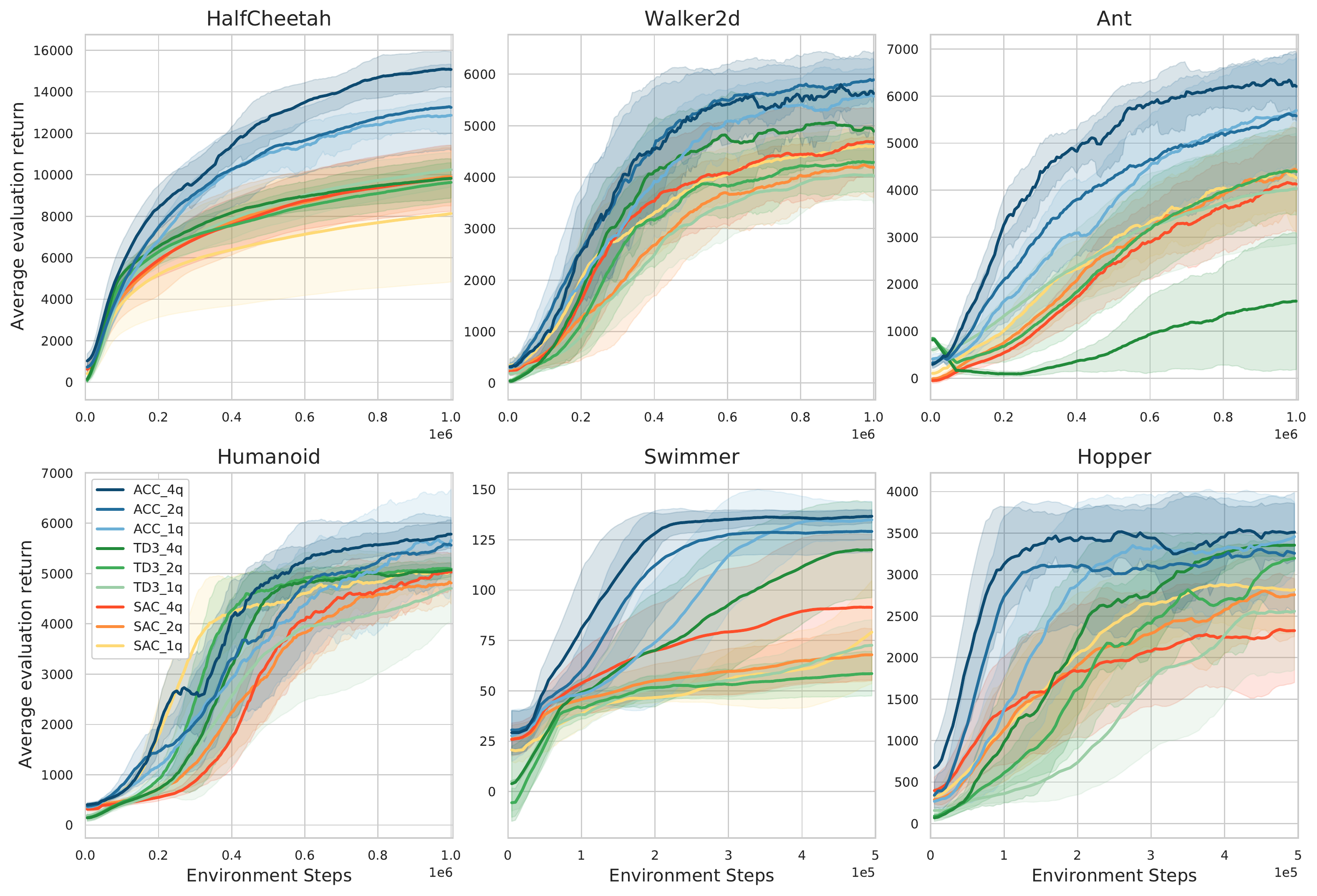} &
    \includegraphics[width=.39\linewidth]{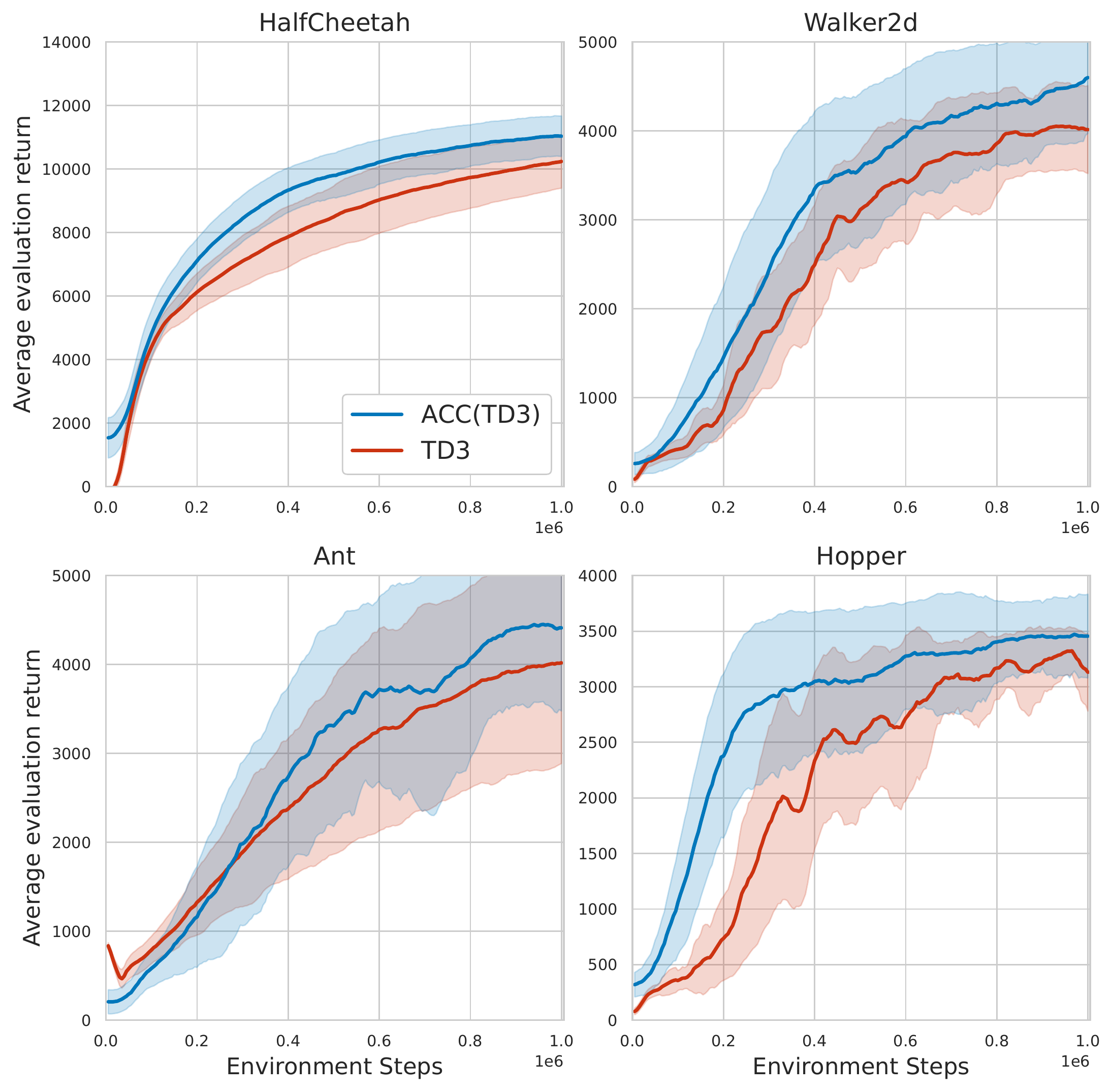} \\
    a) & b) \\
\end{tabular}
\vspace{-0.3cm}
\caption{
The mean $\pm$ standard deviation over $10$ trials. 
\textbf{(a)} Results in the sample efficient regime where tuning of hyperparameters in an inner loop is too costly with different choices for the number of value function updates per environment step.
\textbf{(b)} Results for ACC applied to TD3 compared to pure TD3.}
\label{fig:further_eval}
\vspace{-0.5cm}
\end{figure*}

In principle more critic updates per environment step should make learning faster. However, because of the bootstrapping in the target computation this can easily become unstable.
The problem is that as targets are changing faster, bias can build up easier and divergence becomes more likely.
ACC provides a way to detect upbuilding bias in the TD targets and to correct the bias accordingly.
This motivates to increase the number of gradient updates of the critic.
In TD3, SAC and TQC one critic update is performed per environment step.
We conducted an experiment to study the effect of increasing this rate up to $4$.
ACC using $4$, $2$ and $1$ updates are denoted with ACC\_4q, ACC\_2q and ACC\_1q respectively. ACC\_1q is equal to ACC from the previous experiments. We use the same notation also for TD3 and SAC.

Scaling the number of critic updates by a factor of $4$ increases the computation time by a factor of $4$. But this can be worthwhile in the sample efficient regime, where a huge number of environment interactions is not possible or the interaction cost dominate the computational costs as it is the case when training robots in the real world.
The results in Figure 
\ref{fig:further_eval}a)
show that in the sample efficient regime ACC4q further increases over plain ACC.
ACC4q reaches the final performance of TD3 and SAC in less than a third of the number of steps for five environments and for \textit{Humanoid} in roughly half the number of steps. 
Increasing the number of critic updates for TD3 and SAC shows mixed results, increasing performance for some environments while decreasing it for others. Only ACC benefits from more updates on all environments, which supports the hypothesis that ACC is successful at calibrating the critic estimate.

\subsection{Analysis of ACC}

\begin{figure*}[t]
\footnotesize
\centering 
\begin{tabular}{cc}
    \includegraphics[width=.77\linewidth]{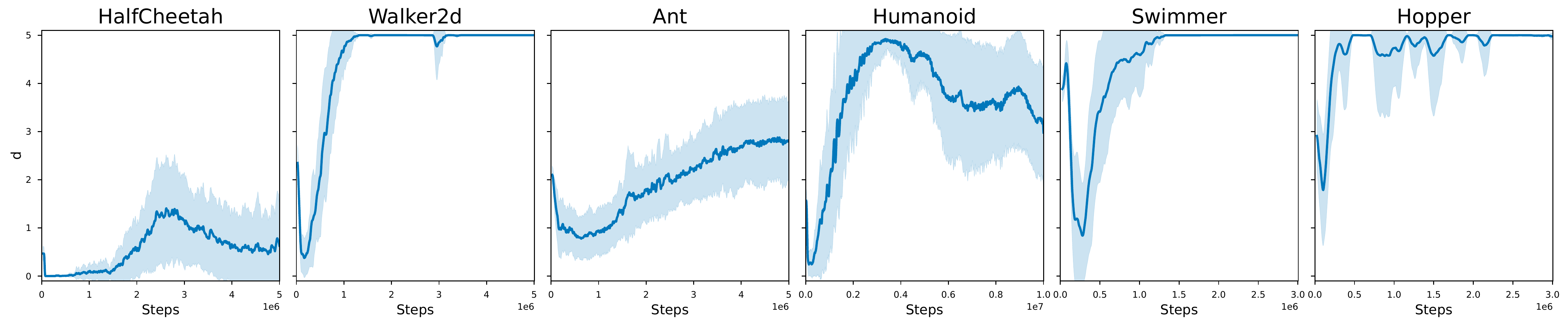} &
    \hspace{-.4cm}\includegraphics[width=.22\linewidth]{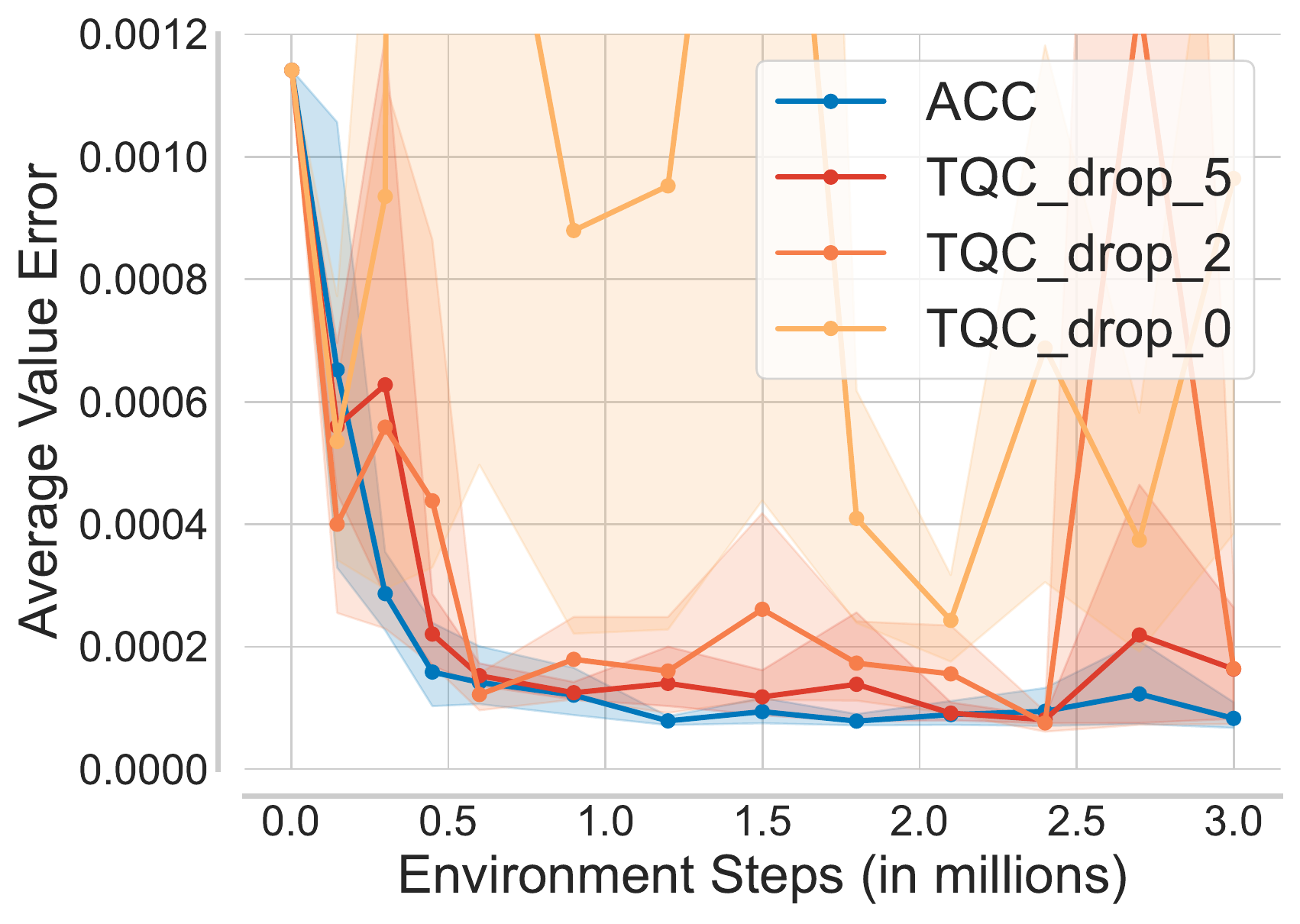} \\
    a) & b) \\
\end{tabular}
\vspace{-0.3cm}
\caption{
\textbf{(a)} Mean (thick line) and standard deviation (shaded area) over 10 trials of the number of dropped targets per network $d = d_{max} - \beta$ in ACC over time for different environments with a uniform filter of size 15.
\textbf{(b)} The normalized absolute error of the value estimate aggregated over the $6$ environments. Shown are the mean with stratified bootstrapped confidence intervals computed from the results of $5$ trials per environment. We used a uniform filter of size $401$ for readability.}
\label{fig:analysis}
\vspace{-0.5cm}
\end{figure*}

To evaluate the effect of ACC on the bias of the value estimate, we analyze the difference between the value estimate and the corresponding observed return when ACC is applied to TQC.
For each state-action pair encountered during exploration, we compute its value estimate at that time and at the end of the episode compare it  with the actual discounted return from that state onwards. Hence, the state-action pair was not used to update the value function at the point when the value estimate has been computed.
If an episode ends because the maximum number of episode time-steps has been reached, which is 1,000 for the considered environments, we ignore the last $100$ state-action pairs. The reason is that in TQC the value estimator is trained to ignore the episode timeout and uses a bootstrapped target also at the end of the episode. 
We normalize for different value scales by computing the absolute error between the value estimate and the observed discounted return and divide that by the absolute value of the discounted return.
Every 1,000 steps, the average over the errors of the last 1,000 state-action pairs is computed.
The aggregated results in Figure 
\ref{fig:analysis}b)
show that averaged over all environments ACC indeed achieves a lower value error than TQC with the a fixed number of dropped atoms $d$.
This supports our hypothesis that the strong performance of ACC applied to TQC indeed stems from better values estimates.

To better understand the hidden training dynamics of ACC we show in Figure
\ref{fig:analysis}a)
how the number of dropped targets per network $d = d_{max} - \beta$ evolves during training.
Interestingly, the relatively low standard deviation indicates a similar behaviour across runs for a specific environment.
However, there are large differences between the environments which indicates that it might not be possible to find a single hyperparameter that works well on a wide variety of different environments.
Further, the experiments shows that the optimal amount of overestimation correction might change over time during the training even on a single environment.

\subsection{Beyond TQC: Improving TD3 with ACC}

To demonstrate the generality of ACC, we additionally applied it to the actor-critic style TD3 algorithm \cite{td3},
which uses two critics. These are initialized differently but trained with the same target value, which is the minimum over the two targets computed from the two critics.
While this successfully prevents the overestimation bias, using the minimum of the two target estimates is very coarse and can instead lead to an underestimation bias.
We applied ACC to TD3 by defining the target for each critic network to be a convex combination between its own target and the minimum over both targets.
Let $Q_i = Q_{\bar{\theta}_i} (s_{t+1}, \pi_{\bar{\phi}} (s_{t+1}) )$, we define the $k$-th critic target
\vspace{-.1cm}
\begin{equation}
\label{eq:td3_target_acc}
    y_k = r + \gamma 
    \Big(   \beta ~ Q_k \nonumber 
     + (1-\beta) \min_{i=1,2} Q_i
    \Big),
\vspace{-.1cm}
\end{equation}
where $\beta \in [0,1]$ is the ACC parameter that is adjusted to balance between under- and overestimation.
The results are displayed in Figure 
\ref{fig:further_eval}b)
and show that ACC also improves the performance of TD3.

\section{Related Work}

\subsection{Overestimation in Reinforcement Learning}

The problem of overestimation in Q-learning with function approximation was introduced by \cite{Thrun+Schwartz:1993}.
For discrete actions the double estimator has been proposed \cite{hasselt2010double} where two Q-functions are learned and one is used to determine the maximizing action, while the other evaluates the Q-function for that action. The Double DQN algorithm extended this to neural networks~\cite{hasselt2016deepdouble}.
However, Zhang \emph{et al.} \cite{weightedQlearning} observed that the double estimator sometimes underestimates the Q-value and propose to use a weighted average of the single and the double estimator as target. This work is similar to ours in the regard that depending on the parameter over- or underestimation could be corrected. A major difference to our algorithm is that the weighting parameter is computed from the maximum and minimum of the estimated Q-value and does not use unbiased rollouts.
Similarly, the weighted estimator
\cite{cini2020deep,d2017estimating} 
estimates the maximum over actions in the TD target as the sum of values weighted by their probability of being the maximum. In continuous action spaces this can be done through Gaussian process regression
\cite{d2017estimating} and for discrete actions via dropout variational inference \cite{cini2020deep}.
Different to ACC the weighting is computed from the same off-policy data used to compute the single quantities while ACC adjusts the weighting parameter $\beta$ in a separate process using the latest on-policy rollouts.
Lv \emph{et al.} \cite{lvSDDQ19} use a similar weighting but suggest a stochastic selection of either the single or double estimator. The probability of choosing one or the other follows a predefined schedule.
Other approaches compute the weighted average of the minimum and maximum over different Q-value estimates \cite{fujimoto2019off,kumarStabilizing19}. However, the weighting parameter is a fixed hyperparameter.
The TD3 algorithm~\cite{td3} uses the minimum over two Q-value estimates as TD target. 
Maxmin Q-learning is another approach for discrete action spaces using an ensemble of Q-functions. For the TD target, first  the minimum of over all Q-functions is computed followed by maximization with respect to the action~\cite{Lan2020Maxmin}. Decreasing the ensemble size increases the estimated targets while increasing the size decreases the targets. Similarly to TQC this provides a way to control the bias in a more fine-grained way; the respective hyperparameter has to be set before the start of the training for each environment, however.
Cetin \emph{et al.}  \cite{cetin2021learning} propose to learn a pessimistic penalty to overcome the overestimation bias.

What sets ACC apart from the previously mentioned works is that unbiased on-policy rollouts are used to adjust a term that controls the bias correction instead of using some predefined heuristic.

\subsection{Combining On- and Off-Policy Learning}
There are many approaches that combine on- and off-policy learning by combining policy gradients with off-policy samples
\cite{degris2012off,NIPS2010_35cf8659,o2016combining}.
In \cite{NIPS2017_IPG} an actor-critic is used where the critic is updated off-policy and the actor is updated with a mixture of policy gradient and Q-gradient. This differs from our work in that we are interested only in better critic estimates through the information of on-policy samples. 
To learn better value estimates by combining on- and off-policy data prior works proposed the use of some form of importance sampling
\cite{NIPS2014_be53ee61,precup2000eligibility}.
In \cite{hausknecht2016policy} the TD target is computed by mixing Monte Carlo samples with the bootstrap estimator.
These methods provide a tradeoff between variance and bias. They differ from our work in using the actual returns directly in the TD targets while we incorporate the returns indirectly via another parameter.
Bhatt \emph{et al.} \cite{bhatt2019crossnorm} propose the use of a mixture of on- and off-policy transitions to generate a feature normalization that can be used in off-policy TD learning. Applied to TD3, learning becomes more stable eliminating the need to use a delayed target network.

\subsection{Hyperparameter Tuning for Reinforcement Learning}

Most algorithms that tune hyperparameters of RL algorithms use many different instances of the environment to find a good setting
\cite{chiang19,falkner18a,jaderberg2017population}. 
There is, however, also work that adjusts a hyperparameter online during training \cite{xu2018meta}. In this work the meta-gradient (i.e., the gradient of the update rule) is used to adjust the discount factor and the length of bootstrapping intervals. However, it would not be straightforward to apply this method to control the bias of the value estimate. Their method also differs from ours in that they do not use a combination of on- and off-policy data.

\section{Conclusion}

We present Adaptively Calibrated Critics (ACC), a general off-policy algorithm that learns a Q-value function with bias calibrated TD targets. 
The bias correction in the targets is determined via a parameter that is adjusted by comparing the current value estimates with the most recently observed on-policy returns.
Our method incorporates information from the unbiased sample returns into the TD targets while keeping the high variance of the samples out. We apply ACC to TQC, a recent off-policy continuous control algorithm that allows fine-grained control of the TD target scale through a hyperparameter tuned per environment.
With ACC, this parameter can automatically be adjusted during training,  obviating the need for extensive tuning.
The strong experimental results suggest that our method provides an efficient and general way to control the bias occurring in TD learning. 

Interesting directions for future research are to evaluate the effectiveness of ACC applied to algorithms that work with discrete action spaces and when learning on a real robot where tuning of hyperparameters is very costly.




\bibliographystyle{IEEEtran}
\bibliography{IEEEabrv,acc_bibliography}


\clearpage

\begin{strip}
\begin{center}
\vspace{-5ex}
\textbf{\LARGE \bf
Adaptively Calibrated Critic Estimates for \\ Deep Reinforcement Learning} \\
\vspace{2ex}

\Large{\bf- Appendix -}\\
\vspace{0.4cm}
\normalsize{Nicolai Dorka\hspace{1cm} Tim Welschehold \hspace{1cm} Joschka Bödecker\hspace{1cm} Wolfram Burgard}\\
\end{center}
\end{strip}

\setcounter{section}{0}
\renewcommand{\thesection}{A.\arabic{section}}
\makeatletter

\section{Proof of Theorem 1}

The estimator  $\qhatpi_{\beta^*}\sa$ was defined via
\begin{equation}
    \beta^* \sa = \argmin_{\beta \in [\bmin, \bmin]} \Bigg| \qhat_\beta \sa - \frac{1}{N} \sum_{i=1}^{N} R_i \sa  \Bigg| .
\end{equation}

To declutter the notation we drop the dependencies on the state-action pairs $\sa$ and the policy $\pi$.
Further we write $\Bar{R} = \frac{1}{N} \sum_{i=1}^{N} R_i $.
First note that the average of symmetrically distributed random variables is still a symmetric distributed random variable and hence 
$\Bar{R} $ is symmetrically distributed.
By assumption $\qhat_{\bmin}$ and $\qhat_{\bmax}$ have the same  distance to the true Q-value which is the mean $Q=\E[\Bar{R}] $, i.e. there is a distance real valued value $d$ such that
$Q = \qhat_{\bmin} +d = \qhat_{\bmax} -d$
Denote the tail probabilty  P($\Bar{R} < \qhat_{\bmin}) = p_t$.  Because of the symmetry and the same distance to the mean we also have that  $P(\Bar{R} > \qhat_{\bmax}) = p_t$.
In the computation of $\E [ \qhat_{\beta^*} ]$  we can differentiate three events.
If $\qhat_{\bmin} \leq \Bar{R} \leq \qhat_{\bmax}$ then $\qhat_{\beta^*} =\Bar{R}$, 
if $\qhat_{\bmin} \geq \Bar{R}$ then $\qhat_{\beta^*} =\qhat_{\bmin}$
and if $\qhat_{\bmin} \geq \Bar{R}$ then $\qhat_{\beta^*} =\qhat_{\bmax}$.
We denote the indicator function with $\idc{A}$, which is equal to $1$ if the event $A$ is true and $0$ otherwise.
Then we get

\begin{align*}
   \E \Big[ \qhat_{\beta^*} \Big] 
   &= \E \Bigg[ \idc{\qhat_{\bmin} \leq \Bar{R} \leq \qhat_{\bmax}} \Bar{R}  \Bigg]  \\
  &~~~~+  \E \Bigg[ \idc{\qhat_{\bmin} \geq \Bar{R} }  \qhat_{\bmin} \Bigg] \\
  &~~~~+  \E \Bigg[ \idc{\qhat_{\bmax} \leq \Bar{R} }  \qhat_{\bmax}  \Big| \Bigg]  \\
   &= (1-2p_t) \cdot \E[\Bar{R}] + p_t \E\Big[\qhat_{\bmin}\Big] + p_t \E\Big[\qhat_{\bmax}\Big] \\
   &= (1-2p_t) Q + p_t \qhat_{\bmin}+ p_t \qhat_{\bmax}\\
   &= (1-2p_t) Q + p_t (Q-d) + p_t (Q+d)\\
   &= (1-2p_t) Q + 2p_t Q + p_t(d-d) \\
   &= Q.
\end{align*}

\section{Pseudocode}
\label{app:pseudocode}

The pseudocode for ACC applied to TQC is in Algorithm \ref{alg:acc_applied_to_tqc}.
As the number of dropped targets per network is given by $d= d_{\max} - \beta $, we state the pseudocode in terms of the parameter $d$ instead of $\beta$.

\begin{algorithm}[t]
   \caption{ACC - Applied to TQC}
   \label{alg:acc_applied_to_tqc}
\begin{algorithmic}
   \STATE {\bfseries Initialize:} $d$ the bias controlling parameter, $\alpha$ the learning rate for $d$, $T_d$ the minimum number of steps between updates to $d$, $T_d^{init}$  the initial steps before $d$ is updated,
   $S_R$ the size from which on episodes are removed from the batch storing the most recent returns, moving average parameter $\tau_d$, $t_d = 0$
   \FOR{$t=1$ {\bfseries to} total number of environment steps}
   \STATE Interact with environment according to $\pi$, store transitions in replay buffer $\mathcal{B}$ and, increment $t_d \pluseq 1$
    \IF{episode ended}
   \STATE Store observed returns $R\sa$ and corresponding state-action pairs $\sa$ in $\mathcal{B}_R$

   \IF{ $t_d >= T_d$ \textbf{and} $t > T_d^{init}$}
   \STATE $C = \sum_{ (s,a, R) \in \mathcal{B}_R} \Big[   Q \sa - R \sa  \Big]$,     $ma = (1-\tau_d) ma + \tau_d C$
   \STATE $d = d + \alpha \frac{C}{ma}$, clip $d$ in interval $[0, d_{max}]$, set $t_d=0$
    \STATE Remove the oldest episodes from $\mathcal{B}_R$ until there are at most $S_R$ left
   \ENDIF
   \ENDIF
   \STATE Sample mini-batch from $\mathcal{B}$
   \STATE Update critic $Q$ as in TQC, where $dN$ (rounded to the next integer) number of targets are dropped from the set of pooled targets 
   \STATE Update policy $\pi$ as in TQC 
  \ENDFOR
\end{algorithmic}
\end{algorithm}

\section{Using Fewer Critic Networks for Faster Runtime}
\label{app:2_nets}

\begin{figure*}[t]
\centering
  \includegraphics[width=.8\linewidth]{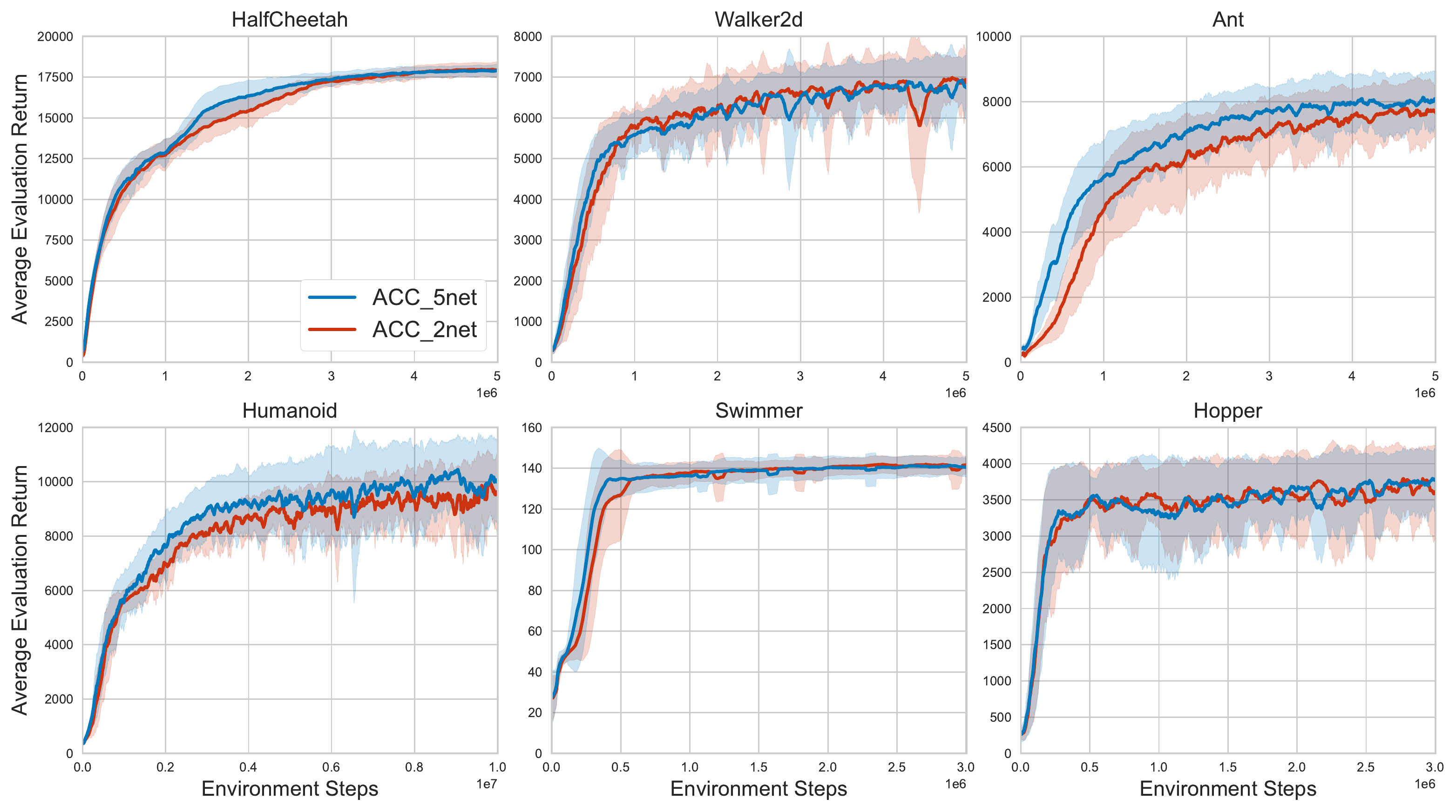}
\caption{The mean $\pm$ standard deviation over $10$ trials. 
Results with different choices for the number of critic networks for each algorithm. }
\label{fig:num_critic_nets}
\end{figure*}
Using $5$ critic networks - the default in TQC - to approximate the value function leads to a high runtime of the algorithm. It is possible to trade off performance against runtime by changing the number of critic networks. We evaluated ACC applied to TQC with $2$ networks and compare it to the standard setting with $5$ networks in Figure \ref{fig:num_critic_nets}. The results show that reducing the number of critic networks to $2$ leads only to a small drop in performance while the runtime is more than $2$ times faster.

\section{Hyperparameters}
\label{app:hyperparameter}

At the beginning of the training we initialize $\beta = 2.5$ and set the step size parameter to $\alpha=0.1$.
After $T_\beta = 1000$ steps since the last update and when the next episode finishes, $\beta$ is updated with a batch that stores the most recent state-action pairs encountered in the environment and their corresponding observed discounted returns. 
The choice of $T_\beta$ was motivated by the fact that the maximum duration of an episode is $1000$ steps for the considered environments.
After every update of $\beta$ the oldest episodes in this stored batch are removed until there are no more than $5000$ state-action pairs left. This means that on average $\beta$ is updated with a batch whose size is a bit over $5000$. 
The updates of $\beta$ are started as soon as $25000$ environment steps as completed and
the moving average parameter in the normalization of the $\beta-$update is set to $0.05$. 
The  first $5000$ environment interactions are generated with a random policy after which learning starts.
Apart from that all hyperparameters are the same as in TQC with $N=5$ critic networks.
In Table \ref{tab:hyperparameter} we list all hyperparameters of ACC applied to TQC.

In the following we also desribe the process of hyperparameter selection.
The range of values $d$ is allowed to take is set to the interval $[0,5]$ as it includes the optimal hyperparameters for TQC from all environments, which are in the set $\{0,2,5\}$. We did not try higher values than $5$.
The initial value for number of dropped targets per network was set to $2.5$ as this value is in the middle of the allowed range and did not evaluated other choices.
The learning rate $\alpha$ of $d$ was set to $0.1$ based on visual inspection of how fast $d$ changes. We  evaluated $\alpha=0.05$ for a small subset of tasks and seeds, but $\alpha=0.1$ gave slightly better results.
$T_d$ was set to $1000$ as the episode length is $1000$ and we did not evaluate other choices.
For $T_d^{init}$ we evaluated the choices $10000$ and $25000$ on a small subset of environments and seeds and did not found a big impact on performance. As $d$ changes very quickly in the beginning we chose $T_d^{init}=25000$.
For $S_R$ we evaluated the choices $1000$ and $5000$ also on a small subset of environments and seeds and found $5000$ to perform slightly better.
We did not tune the moving average parameter and set it to $\tau_d = 0.05$.
For all hyperparameters for which we evaluated more than one choice we do not have definite results as the number of seeds and environments were limited.
The hyperparameters shared with TQC were not changed.
For TD3 and SAC we used the hyperparameters from the respective papers.

\begin{table*}[t]
\caption{Hyperparameters values.}
\label{tab:hyperparameter}
\vskip 0.15in
\begin{center}
\begin{small}
\begin{sc}
\begin{tabular}{lccc}
\toprule
Hyperparameter & \multicolumn{3}{c}{ACC} \\
\midrule
Optimizer & \multicolumn{3}{c}{Adam} \\
Learning rate & \multicolumn{3}{c}{\num{3e-4}} \\
Discount $\gamma$ & \multicolumn{3}{c}{0.99} \\
Replay buffer size & \multicolumn{3}{c}{\num{1e6}} \\
Number of critics $N$ & \multicolumn{3}{c}{5} \\
Number of atoms $M$ & \multicolumn{3}{c}{25}\\
Huber loss parameter  & \multicolumn{3}{c}{1} \\
Number of hidden layers in critic networks & \multicolumn{3}{c}{3}\\
Size of hidden layers in critic networks & \multicolumn{3}{c}{512} \\
Number of hidden layers in policy network & \multicolumn{3}{c}{2} \\
Size of hidden layers in policy network & \multicolumn{3}{c}{256} \\
Minibatch size & \multicolumn{3}{c}{256} \\
Entropy target  &  \multicolumn{3}{c}{$- \dim \mathcal{A}$} \\
Nonlinearity & \multicolumn{3}{c}{ReLU} \\
Target smoothing coefficient  & \multicolumn{3}{c}{0.005} \\
Target updates per critic gradient step & \multicolumn{3}{c}{1} \\
Critic gradient steps per iteration & \multicolumn{3}{c}{1} \\
Actor gradient steps per iteration & \multicolumn{3}{c}{1} \\
Environment steps per iteration & \multicolumn{3}{c}{1} \\
\midrule
Initial value for number of dropped targets per network & \multicolumn{3}{c}{2.5} \\
Maximum value for $d$ denoted $d_{\max}$ & \multicolumn{3}{c}{5} \\
Minimum value for $d$ denoted $d_{\min}$ & \multicolumn{3}{c}{0} \\
Learning rate for $d$ denoted $\alpha$ & \multicolumn{3}{c}{0.1} \\
Minimum number of steps between updates to $d$ denoted $T_d$ & \multicolumn{3}{c}{1000} \\
Initial number of steps before $d$ is updated denoted $T_d^{init}$ & \multicolumn{3}{c}{25000} \\
Limiting size for batch used to update $d$ denoted $S_R$ & \multicolumn{3}{c}{5000} \\
Moving average parameter $\tau_d$ & \multicolumn{3}{c}{0.05} \\
\midrule
\midrule
Hyperparameter in Sample Efficient Experiment & ACC\_1q & ACC\_2q & ACC\_4q \\
\midrule
Critic gradient steps per iteration & 1 & 2 & 4 \\
Actor gradient steps per iteration & 1 & 1 & 1 \\
Target updates per critic gradient step & 1 & 1 & 1\\
\bottomrule
\end{tabular}
\end{sc}
\end{small}
\end{center}
\vskip -0.1in
\end{table*}

\section{Potential Limitations}
One limitation of our work is that ACC can not be applied in the offline RL setting, as ACC also uses on-policy data.
Furthermore, in the stated form ACC relies on the episodic RL setting. However, we believe that ACC could potentially be adapted to that setting. 
It is also not entirely clear how the algorithm would perform in the terminal reward setting, where a reward of for example $1$ is given upon successful completion of a specific task. While we do not have experiments for such environments we imagine that the positive effect of ACC could diminish as the true Q-values of states closer to the start of the episode are almost zero because of the discounting.

\section{Analysis of the ACC Parameter}
\label{app:analysis_acc_parameter}

\begin{figure*}[b]
  \includegraphics[width=1\linewidth]{images/analysis/visualize_beta_all_envs.pdf}
  \includegraphics[width=1\linewidth]{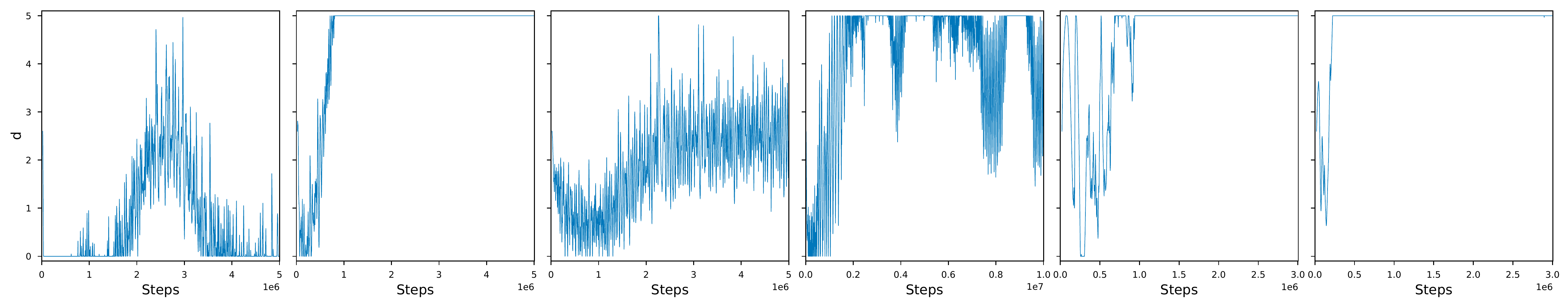}
\caption{Development of the number of dropped targets per network $d=d_{max}-\beta$ in ACC over time for different environments. The top row shows the mean (thick line) and standard deviation (shaded area) over the $10$ trials where for readability a uniform filter of size $15$ is used.
The bottom row shows the unfiltered development for one of the seeds.}
\label{fig:num_dropped_targets_all_envs}
\end{figure*}

To better understand the hidden training dynamics of ACC we show in 
Figure \ref{fig:num_dropped_targets_all_envs}
how the number of dropped targets per network $d=d_{max}-\beta$ evolves during training. 
To do so we plotted $d$ after every $5000$ steps during the training of ACC.
From the top row the first observation is that per environment the results are similar over the $10$ seeds as can be seen from the relatively low standard deviation. We show the single runs for all seeds in the appendix to further support this observation.
However, there are large differences between the environments which supports the argument that it might not be possible to find a single hyperparameter that works well on a wide variety of different environments.
Another point that becomes clear from the plots is that the optimal amount of overestimation correction might change over time during the training even on a single environment.

In the bottom row of Figure \ref{fig:num_dropped_targets_all_envs} we plotted the evolution of $d$ for one of the $10$ trials in order to shed light on the actual training mechanics of a single run without lost information due to averaging.
For each environment there is a trend but $d$ is also fluctuating to a certain degree.
While this shows that the initial value of $d$ is not very important as the value quickly changes, this also highlights another interesting aspect of ACC. 
The rollouts give highly fluctuating returns. The parameter $d=d_{max}-\beta$ is changing more slowly and picks up the trend. So a lot of the variance of the returns is filtered out in ACC by incorporating on-policy samples via the detour over $\beta$.
This leads to relatively stable TD targets computed from $\qbeta$ while an upbuilding under- or overestimation is prevented as $\beta$ picks up the trend. On the other hand, if $\beta$ would change too slowly the upbuilding of the bias might not be stopped.





\end{document}